\documentclass[times,10pt,twocolumn]{article}

\usepackage{latex8}
\usepackage{times}
\usepackage{booktabs} 
\usepackage{float}
\usepackage{amsmath}
\usepackage{epsfig}
\DeclareMathOperator*{\argminA}{arg\,min}
\usepackage{subcaption}
\usepackage{multirow}
\usepackage{graphicx}
\usepackage{amssymb}
\usepackage{caption}
\usepackage[capitalize]{cleveref}

\begin{document}

\title{Real-Time Super-Resolution for Real-World Images on Mobile Devices}

\author{
\large{Jie Cai, Zibo Meng, Jiaming Ding, and Chiu Man Ho}\\
\small{OPPO US Research Center, Palo Alto, CA, US}\\
\footnotesize{ \{jie.cai, zibo.meng, jiaming.ding, chiuman\}@oppo.com}
}

\maketitle
\thispagestyle{empty}

\begin{abstract}
Image Super-Resolution (ISR), which aims at recovering High-Resolution (HR) images from the corresponding Low-Resolution (LR) counterparts.
Although recent progress in ISR has been remarkable.
However, they are way too computationally intensive to be deployed on edge devices, since most of the recent approaches are deep learning based. 
Besides, these methods always fail in real-world scenes, since most of them adopt a simple fixed ``ideal" bicubic downsampling kernel from high-quality images to construct LR/HR training pairs which may lose track of frequency-related details.
In this work, an approach for real-time ISR on mobile devices is presented, which is able to deal with a wide range of degradations in the real-world scenarios.
Extensive experiments on traditional super-resolution datasets (Set5, Set14, BSD100, Urban100, Manga109, DIV2K) and real-world images with a variety of degradations demonstrate that our method outperforms the state-of-art methods, resulting in higher PSNR and SSIM, lower noise and better visual quality. 
Most importantly, our method achieves real-time performance on mobile or edge devices.
\end{abstract}

\section{INTRODUCTION}

Image super-resolution~\cite{yang2019deep,wang2020deep}, which stands for the process of restoring HR images from LR ones, has been studied as an important class of image processing and computer vision techniques for decades. 
ISR is used in a wild range of real-world applications, such as medical image~\cite{huang2017simultaneous}, surveillance~\cite{zhang2010super} and oceanography~\cite{ducournau2016deep}. 
However, ISR is very challenging and inherently ill-posed problem since there always exists multiple possible HR solutions corresponding to the same LR image.

With the rapid development of deep learning techniques, deep learning based ISR models~\cite{dong2014learning,ledig2017photo,wang2018esrgan} have been actively developed and achieved compelling performance. 
Specifically, the achievement ranges from the promising super-resolution methods using Convolutional Neural Networks (CNNs) (e.g., SRCNN~\cite{dong2014learning}) to recent Generative Adversarial Nets (GANs)~\cite{goodfellow2014generative} based approaches (e.g., SRGAN~\cite{ledig2017photo}, ESRGAN~\cite{wang2018esrgan}).
In general, the design of deep learning based super-resolution methods differ from each other in the following major aspects: (1) different types of network structures; (2) different types of loss functions; (3) different types of learning strategies.

However, the biggest limitation of the above methods is that they primarily aim at achieving highest PSNR and SSIM while not optimize for computational efficiency in terms of parameters and FLOPs. Specifically, they contain millions of parameters, more than 600G FLOPS, and cost several seconds to inference single LR frame on GPU. Therefore, it is impossible to deploy them on mobile devices. For our proposed model, it only contains 50k paraments, 20G FLOPs, and takes about 20ms to inference one frame on mobile NPU. Moreover, we need to consider mobile-related constraints, e.g., power consumption, RAM amount restriction, and CNN operations compatibility.

Besides, unlike conventional ISP task that can be easily solved through supervised learning since the degradations only include bicubic downsample and Gaussian blur, the degradations in real-world images are complex and the performance usually degrades dramatically. For example, when we take photos using smartphones, the photos may contain several degradations, such as camera sensor noise, artifacts, and JPEG compression. When we upload photos to Internet, which may further introduce unpredictable compression and noises. When we send photos via social media APP, which may cause signal transmission compression and downsampling. Therefore, the real-world degradations are usually too complex to be modeled with a simple combination of multiple degradations.

Although several degradation models take additional factors into consideration, such as different kinds of blur, they are still not effective enough to cover the diverse degradations of real-world images/videos. To address this issue, we propose a complex but practical degradation pipeline that consists of blur, noise, compression loss, downsampling, and random shuffle data augmentation strategy. 
Specifically, the blur is approximated by two convolutions with isotropic and anisotropic Gaussian kernels, 2D sinc filter blur, and blind kernel estimation. The noise is synthesized by adding Gaussian noise and Poisson noise. Besides, we explicitly estimate noise from the LR images. The compression loss is achieved by adjusting JPEG compression with different quality factors. The downsampling is randomly chosen from nearest, bilinear, bicubic interpolations, and real-world collected LR/HR pairs. Moreover, we involve a degradation shuffle strategy to expand degradation space. In addition to augmenting degradation space on LR domain, we further show a training trick to visually improve the image sharpness, while not introducing visible artifacts. 

In summary, our major contributions are:

\begin{enumerate}
\item[-] Proposing a novel real-time ISR model for mobile devices and could achieve 50 fps for VSR applications;
\item[-] Proposing a novel data degradation pipeline for ISR, which aims to recover LR data from real scenes.
\end{enumerate}

Experiments show that InnoPeak\_mobileSR yields better perceptual quality and achieves comparable or better performance compared with the baseline models and the state-of-the-art methods.

\section{RELATED WORK}

\subsection{Image Super-Resolution}
The pioneer work was achieved by Dong et al.~\cite{dong2014learning}, who developed SRCNN by introducing a three-layer CNN for ISR. 
Kim et al. increased the CNN depth to 20 in VDSR~\cite{kim2016accurate} and DRCN~\cite{kim2016deeply} and achieved notable improvements over SRCNN later on.
Lim et al.~\cite{lim2017enhanced} proposed a very deep network MDSR and a very wide network EDSR by utilizing multiple residual blocks.
Tai et al.~\cite{tai2017image} developed a very deep yet concise Deep Recursive Residual Network (DRRN) which consists up to 52 convolutional layers.
Huang et al.~\cite{huang2017densely} proposed Dense Convolutional Network (DenseNet), which connects each layer to every other layer in a feed-forward fashion. 
Tong et al.~\cite{tong2017image} and Zhang et al.~\cite{zhang2018residual} jointly deployed residual block and dense block in very deep networks, which provide an effective way to combine the low-level features and high-level features and further boost the reconstruction performance for ISR.

\subsection{Real-World Image Super-Resolution}
Most recent proposed approaches rely on paired LR/HR images to train the deep network in a fully supervised manner. However, such image pairs are not often available in most real-world applications. 
The AIM 2019 Challenge on Real-World Image Super-Resolution~\cite{AIM2019RWSRchallenge} and the NTIRE 2020 Challenge on Real-World Image Super-Resolution~\cite{NTIRE2020RWSRchallenge} aim to stimulate research in the direction of real-world ISR, i.e., no paired reference HR images are provided for training. 
Fritsche et al.~\cite{fritsche2019frequency} proposed DSGAN which trained in an unsupervised manner on HR images. Specifically, they first generated low-resolution images with the same characteristics as the original images; and then utilized the generated data to train a super-resolution model, which improves the performance on real-world images. 
Ji et al.~\cite{ji2020real} proposed a novel degradation framework RealSR based on kernel estimation and noise injection. The method is the winner of NTIRE 2020 Challenge and achieves significant improvement than other competitors.
Cai et al.~\cite{cai2020residual} proposed a novel RCA-GAN which consists of residual channel attention module and GAN to restore the finer texture details for ISR under real-world settings.

\subsection{Lightweight Image Super-Resolution}
While many challenges and works~\cite{zhang2020aim} targeted at efficient deep learning models have been proposed recently, the evaluation of the obtained solutions is generally performed on desktop CPUs and GPUs, making the developed solutions not practical due to the restricted
amount of edge device RAM, limited and not supporting deep learning operators, and power consumption limitation.
The MAI 2021 Challenge on ISR~\cite{ignatov2021real} aims to stimulate research in the direction of deeping learning solutions are developed for mobile devices.
Ayazoglu et al.~\cite{ayazoglu2021extremely} proposed a hardware (Synaptics Dolphin NPU) limitation aware, extremely lightweight quantization robust real-time super resolution network (XLSR). 
Du et al.~\cite{du2021anchor} aim at designing efficient architecture for 8-bit quantization and deploy it on mobile device. Specifically, they proposed anchor-based plain
net (ABPN) and adopted Quantization Aware Training (QAT) strategy to further boost the performance. 

\section{METHODOLOGY}
In this section, we first introduce the proposed real-time InnoPeak\_mobileSR for mobile devices.
Then, we show blind super-resolution and the proposed classical degradation model for real-world images. 
Finally, we describe the overall loss function of the InnoPeak\_mobileSR.

\begin{figure}[th]
   \centering
   \includegraphics[width=0.5\textwidth]{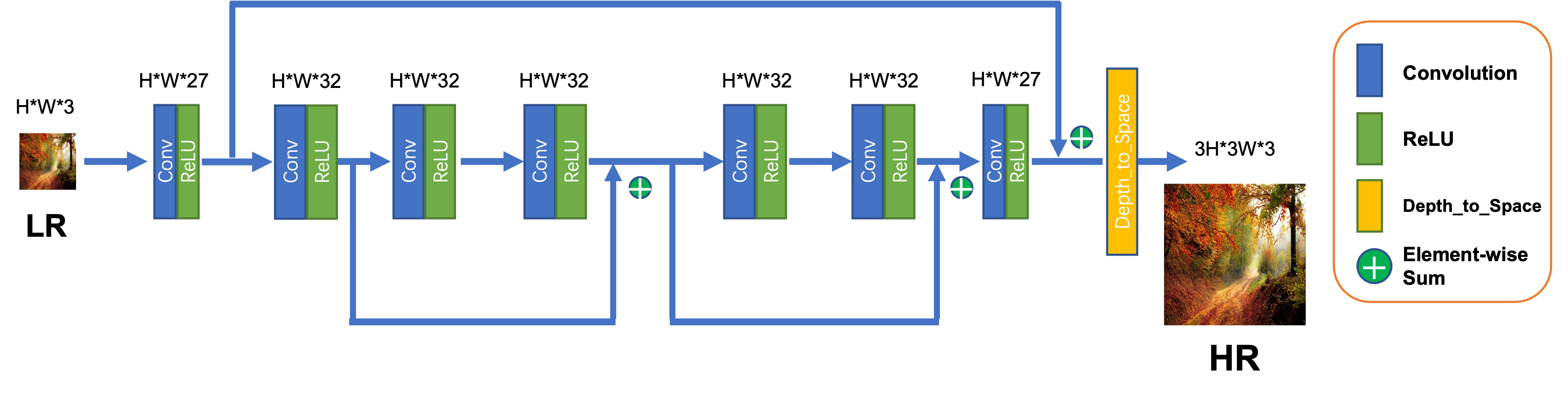}
   \caption{\small{An illustration of the proposed InnoPeak\_mobileSR for mobile devices. Best viewed in color.}}
   \label{fig:model}
\end{figure}

\subsection{InnoPeak\_mobileSR}

Our goal is to design a real-time ISR model for mobile devices. There are several constrains that prevent CNN deployment on mobile devices: a restricted amount of RAM, many common CNN operators are not supported, limited FLOPs, and power consumption requirement for mobile device. Thus, it requires a careful network architecture design. Moreover, we also utilize full 8-bit QAT strategy and design neural architecture using hardware friendly operations.

First, we need to figure out the set of portable meta-operators and time-consuming meta-operators. Inspired by~\cite{ayazoglu2021extremely,du2021anchor,ignatov2021real}, convolution, transposed convolution, concatenation, element-wise summation, depth\_to\_space, space\_to\_depth, and ReLU are selected to form InnoPeak\_mobileSR. Specifically, these operations can be divided into four categories: tensor operator nodes (Concatenation and Summation), convolution nodes (Convolution and Transposed Convolution), activation nodes (ReLU), and resize nodes (Convolution, Transposed Convolution, depth\_to\_space and space\_to\_depth).

The whole architecture is shown in Fig.~\ref{fig:model}. The proposed architecture mainly consists of four parts: shallow feature extraction part which transfers LR image to feature space, deep feature extraction part which learns high-level information and restores details (edges, textures, et al.), skip connection part, and reconstruction part which maps feature space to HR image.

\subsection{Blind Super-Resolution}

Recent state-of-the-art ISR methods have achieved impressive performance on ideal datasets that only contains traditional bicubic downsmaple. However, these methods always fail in real-world ISR datasets, e.g., DIV2K\_Real and DPED, with a variety of degradations. Inspired by~\cite{ji2020real,NTIRE2020RWSRchallenge}, we utilize the novel degradation framework for real-world images by estimating real noise distributions and various blur kernels. Specifically, we decompose the real-world super-resolution problem into two sequential steps: (1) estimating noise and blur kernel from LR images and (2) restoring SR images based on estimated kernel and noise.

\subsubsection{Noise Injection}

We explicitly inject estimated noises into the downsampled HR images to generate realistic LR images. In order to make the degraded LR image have a similar noise distribution to the LR image, we directly collect noise patches from the LR dataset. Inspired by~\cite{ji2020real,NTIRE2020RWSRchallenge}, we use the filtering rule to collect noise patches with their variance in a certain range:

\begin{equation} \label{eq:noise_injection}
 \begin{aligned}
\sigma(n_{i})<v,
 \end{aligned}
\end{equation}
where $\sigma(\cdot)$ denotes the function to calculate variance in the certain patch $n_{i}$, and $v$ is the max value of the pre-defined variance.

\subsubsection{Kernel Estimation}

We use a kernel estimation algorithm to explicitly estimate kernels from real LR images. Inspired by KernelGAN~\cite{bell2019blind}, we utilize a similar kernel estimation method. The generator of KernelGAN is a linear model without any activation layers, therefore the parameters of all layers can be combined into a fixed kernel. The estimated kernel needs to meet the following constraints:

\begin{equation} \label{eq:kernel_GAN}
 \begin{aligned}
\argminA_{k} & \| (I_{src} \times k) \downarrow_{s} - I_{src} \downarrow_{s} \|_1 + |1- \sum k_{i,j}| 
\\ &
+ |\sum k_{i,j} \cdot m_{i,j}| + |1- D((I_{src} \times k) \downarrow_{s}|
 \end{aligned}
\end{equation}

$ (I_{src} \times k) \downarrow_{s} $ is downsampled LR images convolved with kernel $ k $, and $ I_{src} \downarrow_{s} $ is downsampled image with ideal kernel, therefore to minimize this error is to encourage the downsampled image to preserve important low-frequency information of the source image. Besides, the second term of the above formula is to constrain the summation of kernel $k$ to 1, and the third term is to penalty boundaries of $k$. Finally, the discriminator $D(\cdot)$ is to ensure the consistency of source domain.

\subsection{A Practical Degradation Model}

\begin{figure}[th]
   \centering
   \includegraphics[width=0.5\textwidth]{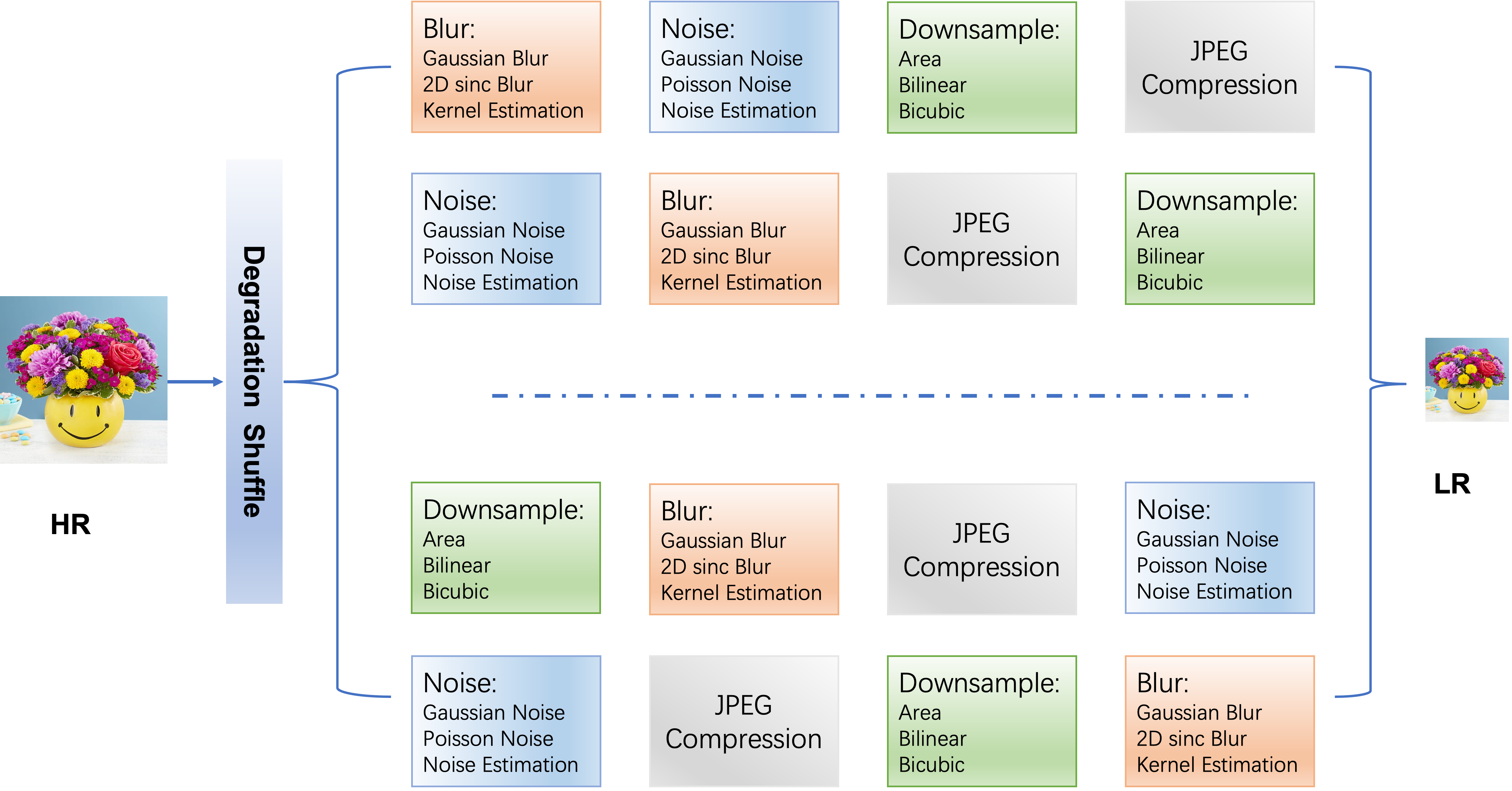}
   \caption{\small{Overview of the pure synthetic data generation pipeline. Best viewed in color.}}
   \label{fig:data}
\end{figure}

Though noise injection and kernel estimation have been shown great improvement for blind ISR to restore LR images with complex unknown degradations, they are still far from addressing general real-world LR images with a variety of degradations.
In this work, we further design a practical degradation pipeline, which is trained with pure synthetic data, to deal with complex real-world settings. 
Fig.~\ref{fig:data} shows the detailed degradation pipeline including different choices of blur, noise, dowmsampling, JPEG compression and the proposed random shuffle strategy.

\noindent  \emph{\textbf{Blur:}} We typically model blur degradation as a convolution with a linear blur kernel. Isotropic and anisotropic Gaussian filters are typical choices. To avoid ringing and overshoot artifacts, we  further employ the 2D sinc filter which is an idealized filter that cuts off high-frequency information. Besides, we also use the above kernel estimation algorithm to explicitly estimate SR kernels from real images.

\noindent  \emph{\textbf{Noise:}} First, we consider two popular noises, i.e., Gaussian and Poisson. Besides, we also utilize the above noise injection method.

\noindent  \emph{\textbf{Downsampling}} is a basic operation for synthesizing LR images. There are several resize algorithms, e.g., area resize, bilinear interpolation, and bicubic interpolation. 

\noindent  \emph{\textbf{JPEG Compression}} is a common compression loss technique for digital image. The degree of compression is determined by the quality factor which is an integer in the range [0, 100]. The quality factor 100 means lower compression and higher quality, and vice versa.

\noindent  \emph{\textbf{Random Shuffle:}} In order to cover the diverse degradations of real images, we design a more complex but practical degradation pipeline that consists of randomly shuffled blur, noise, downsampling, and JPEG compression.

\noindent  \emph{\textbf{Sharpen ground-truth images during training.}} We further show a training trick to improve the visual sharpness, while not introducing unpleasant artifacts. A traditional way of sharpening images is to employ a post-processing sharping algorithm. However, this algorithm often introduces overshoot artifacts and unpleasant side effects. We empirically find that sharpening ground-truth images during training could achieve a better balance between sharpness and overshoot artifact suppression.

\subsection{Loss Function}

The overall loss function of the InnoPeak\_mobileSR is defined as:
\begin{equation} \label{eq:joint_IF-GAN}
 \begin{aligned}
\mathcal{L} = \lambda_{1} \cdot \mathcal{L}_{L1} + \lambda_{2} \cdot \mathcal{L}_{SSIM}   + \lambda_{3} \cdot \mathcal{L}_{VGG}
 \end{aligned}
\end{equation}
where the hyperparameters $\lambda_{1}=1$, $\lambda_{2}=0.3$, $\lambda_{3}=0.3$ are used to balance the three terms.

\subsubsection{Pixel Loss}
$\mathcal{L}_{L1}$ learns to capture the true data distribution to generate realistic images that are similar to the images sampled from the true data distribution. 
\begin{equation} \label{eq:L1}
\mathcal{L}_{L1}={\mathbb{E}} [\lVert \mathbf{I}_{_{HR}} - \mathbf{I}_{_{SR}} \rVert_{1}]
\end{equation}

The pixel-wise $L1$ loss constrains the generated $\mathbf{I}_{_{SR}} $ to be close enough to the ground truth $\mathbf{I}_{_{HR}} $ on the pixel values.
Comparing with $L_1$ loss, the $L_2$ loss penalizes larger errors but is more tolerant to small errors, and thus often results in too smooth results. In practice, the $L_1$ loss shows better performance and convergence over $L_2$ loss.
Since the definition of PSNR is highly correlated with pixel-wise difference thus minimizing pixel loss directly maximize PSNR, the pixel loss gradual becomes the most widely used loss function.

\subsubsection{SSIM Loss}
Considering that the human visual system is highly adapted to extract image structures, the structural similarity index (SSIM)~\cite{wang2004image} is proposed for measuring the structural similarity between images based on independent comparisons in terms of luminance, contrast, and structures.

\subsubsection{Perceptual Loss}
However, since the pixel loss actually doesn't take image quality (e.g., perceptual quality~\cite{johnson2016perceptual}, image texture~\cite{sajjadi2017enhancenet}) into consideration, the outputs often lack high-frequency details and are perceptually unsatisfying with oversmooth textures.
In order to further improve perceptual quality of images, the perceptual loss is introduced into super-resolution~\cite{johnson2016perceptual,dosovitskiy2016generating}. 
Specifically, it measures the semantic differences between images using a pre-trained image classification network $\phi$, e.g., VGG-19. 
Denoting the extracted high-level representations on $l$-th layer as $\phi^{l}(I)$, the perceptual loss is indicated as the $L1$ distance between high-level representations of two images:

\begin{equation}  \label{eq:VGG}
\mathcal{L}_{VGG} ( \mathbf{\emph{G}} ) = \mathbb{E} [\lVert \phi^{l}({\mathbf{I}_{_{HR}}}) - \phi^{l}({\mathbf{I}_{_{SR}}}) \rVert_{1}]
\end{equation}



\section{EXPERIMENTS}
\begin{table*}[th!]
  \begin{center}
  	\caption{\small{Comparison of the proposed method with public benchmark floating-point methods.}}
	\scalebox{0.9}{
    \begin{tabular}{c|c|c|c|c|c|c|c}
    \hline
    Method              &    Bicubic    &   FSRCNN     &   ESPCN    &    IMDN    &   VDSR     &    EDSR      &  Ours         \\
    \hline
    \hline
    Parameters          &    --	       &   25k        &   31k      &   500k     &    668k    &    43M       &  53.15k        \\
    \hline
    Set5                &    30.41	   &   33.16      &   33.13    &   34.36    &    33.66   &    34.65     &  33.91       \\
    Set14               &    27.55	   &   29.43      &   29.43    &   30.32    &    29.77   &    30.52     &  30.17        \\
    BSD100              &    27.22	   &   28.60      &   28.50    &   29.09    &    28.82   &    29.25     &  29.14        \\
	Urban100            &    24.47	   &   26.48      &   26.41    &   28.17    &    27.14   &    28.80     &  27.36        \\
	Manga109            &    26.99	   &   30.98      &   30.79    &   33.61    &    32.01   &    32.10     &  31.11        \\
	\hline
	DIV2K Val           &    28.22	   &   29.67      &   29.54    &   30.90    &    30.09   &    31.26     &  30.99        \\
    \hline
    \hline
    \end{tabular}}
    \label{tab:results_three}
  \end{center}
\end{table*}

\begin{table}[th!]
  \begin{center}
  	\caption{\small{Comparison of the proposed method trained under float32, PTQ, and QAT.}}
	\scalebox{0.9}{
    \begin{tabular}{c|c|c|c|c|c|c|c}
    \hline
    Method            &   FSRCNN     &   XLSR    & ESPCN      &       ABRL     &    Ours              \\
    \hline
    \hline
    Parameters        &   25k        &   22k      &   31k     &      43k     &    53.15k           \\
    \hline
    float32           &   29.67      &   30.10    &  29.54    &    30.22     &    30.99            \\
    PTQ               &   21.95      &   29.82    &  23.93    &    30.09     &    30.22             \\
    QAT               &   --         &   --       &   --      &    30.15     &    30.79            \\
    \hline
    \hline
    \end{tabular}}
    \label{tab:QAT}
  \end{center}
\end{table}

To evaluate the proposed InnoPeak\_mobileSR, experiments have been conducted on six public ISR datasets (Set5~\cite{bevilacqua2012low}, Set14~\cite{zeyde2010single}, BSD100~\cite{martin2001database}, Urban100~\cite{huang2015single}, Manga109~\cite{matsui2017sketch}, and DIV2K~\cite{agustsson2017ntire}) and then performed on  low-quality images collected from real-world scenes.

\subsection{Training Details}

All experiments are performed with a scale factor of 3 between LR and HR. This corresponds to a 9 expansion in image pixels. 
DIV2K consists of 800 high quality training images, 100 validation images, and 100 test images. Since the test images are not public released, we report our results on DIV2K validation. 
For experiments on standard benchmark datasets (Set5, Set14, BSD100, Urban100, Manga109, and DIV2K validation), we all use 800 high-quality DIV2K training images. 
For a fair comparison with existing work, we perform Y channel PSNR evaluation for Set5, Set14, BSD100, Manga109, and Urban100; and RGB channel PSNR evaluation for DIV2K validation.
For experiments on real-world images, we collect 2,000 high-quality images from Internet and applied the proposed image degradation pipeline (Fig.~\ref{fig:data}) to generate LR/HR training pairs.

In this work, the total number of epoch is 200. The learning rate $\alpha$ is 1e-4 and reduced by a factor of 0.8 every 10 epochs. 
Adam optimizer with a mini-batch size of 50, $\beta1=0.9$, and $\beta2=0.99$, is used for training the baseline models (float32 and PTQ) and InnoPeak\_mobileSR (QAT).

\subsection{Experimental Results on Standard Benchmark Datasets}

As shown in Table~\ref{tab:results_three}, the proposed InnoPeak\_mobileSR not only beats FSRCNN and ESPCN in terms of PSNR score, but also achieves comparison results with IMDN, VDSR, and EDSR which contain much more parameters.
Experiments have been conducted on DIV2K dataset to show the effect of uint8 quantization, e.g., PTQ (Post-Training Quantization) and QAT (Quantization-Aware Training). As shown in Table~\ref{tab:QAT}, the proposed InnoPeak\_mobileSR achieves the best performance among the float32 and uint8 quantization models.  The QAT network only lose 0.2dB compared with its floating-point version. Most importantly, the proposed uint8 InnoPeak\_mobileSR model could run 20ms on Find X3 Pro NPU which is about 50 fps (frames per second).

\subsection{Visualization Study in Real-World Scenes}

\begin{figure}[th]
   \centering
   \includegraphics[width=0.5\textwidth]{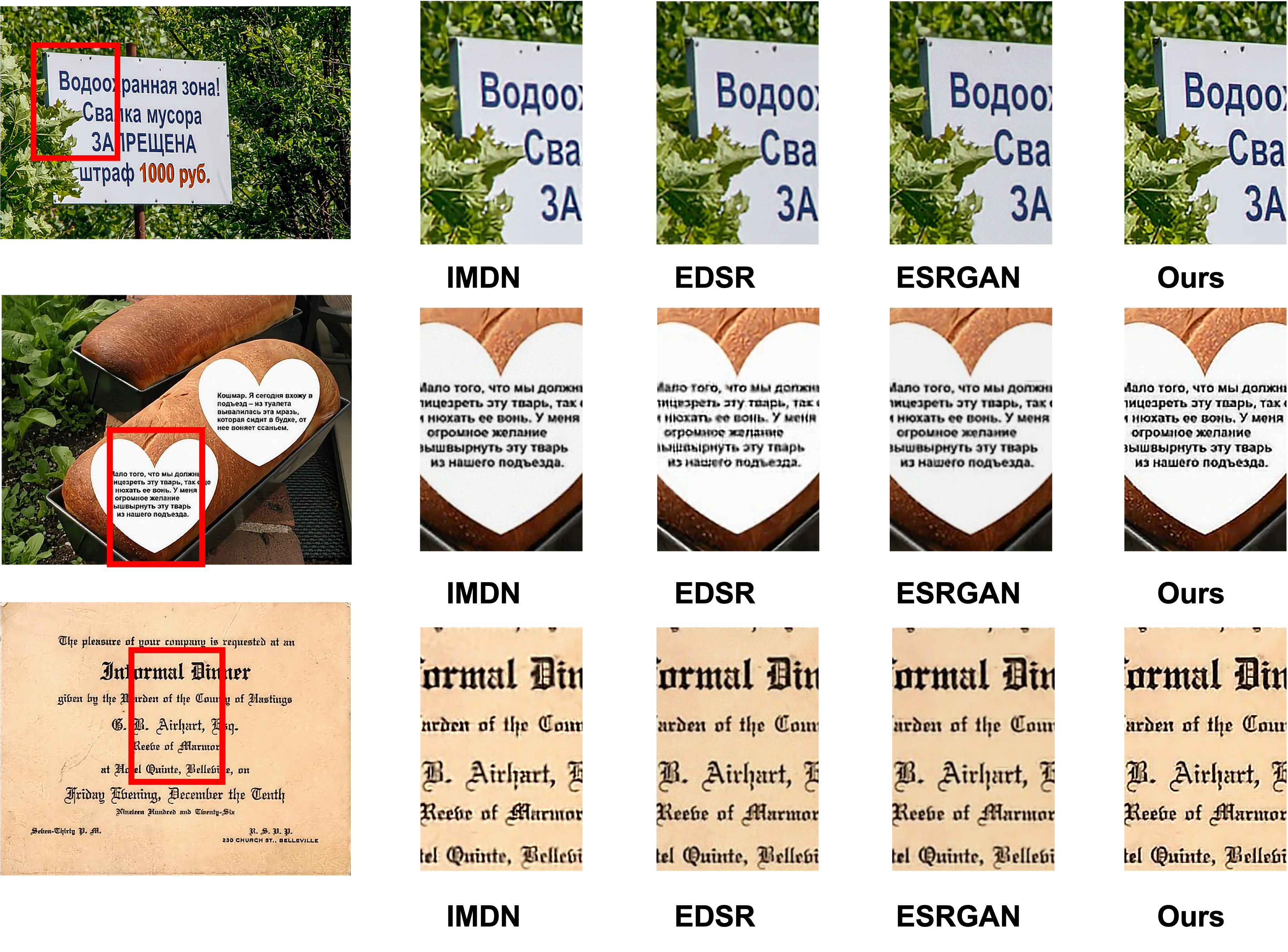}
   \caption{\small{Qualitative comparisons with state-of-the-art methods for real-world images. The proposed degradation pipeline outperforms previous approaches in removing noise and artifact; and restoring finer texture details. Moreover, InnoPeak\_mobileSR model trained under sharpening ground-truths trick can further boost visual sharpness. Best viewed in color.}}
   \label{fig:vil}
\end{figure}

As shown in Fig.~\ref{fig:vil}, we show visual comparison for (1) IMDN model, (2) EDSR model, (3) ESRGAN model, and (4) InnoPeak\_mobileSR model (Fig.~\ref{fig:model}) trained under the proposed degradation pipeline. Note that, the InnoPeak\_mobileSR model only contains 53k parameters compared with IMDN (500k), EDSR (43M), and ESRGAN (16.7M) which demonstrates the effeteness of the proposed degradation pipeline. The lightweight InnoPeak\_mobileSR model could run 50 fps on mobile devices which achieves real-time video super-resolution standards.

\section{CONCLUSION}
In this paper, we propose a real-time ISR model which is efficient in terms of runtime, parameters, FLOPs, and power consumption for mobile devices. It could run 50 fps on most mobile hardwares. Furthermore, the proposed method outperforms state-of-the-art methods in most of public datasets in terms of PSNR. The proposed architecture is very robust to uint8 quantization and only 0.2dB PSNR drop when compared with its float32 model on DIV2K validation.
Besides, we also propose a complex but practical solution for training image/video SR models using pure synthetic data. Extensive comparisons have shown its superior visual performance than prior works on various real-world images.

{
\small
\bibliographystyle{latex8}
\bibliography{../../../reference/abbrev_short,../../../reference/my_paper,../../../reference/my_paper_OPPO,../../../reference/Deep_Learning/deep_learning_toolbox,../../../reference/Deep_Learning/deep_learning_technique,../../../reference/Deep_Learning/CNN,../../../reference/Deep_Learning/GAN,../../../reference/Image_Super_Resolution/NTIRE2020,../../../reference/Image_Super_Resolution/Image_SR_papers,../../../reference/Image_Super_Resolution/Image_SR_application,../../../reference/Image_Super_Resolution/Image_SR_measurement,../../../reference/Image_Super_Resolution/MAI2021,../../../bibliography/ty-literature_database,../../../reference/my_paper_OPPO}

\begin{thebibliography}{10}\setlength{\itemsep}{-1ex}\small

\bibitem{agustsson2017ntire}
E.~Agustsson and R.~Timofte.
\newblock Ntire 2017 challenge on single image super-resolution: Dataset and
  study.
\newblock In {\em CVPR Workshops}, pages 126--135, 2017.

\bibitem{ayazoglu2021extremely}
M.~Ayazoglu.
\newblock Extremely lightweight quantization robust real-time single-image
  super resolution for mobile devices.
\newblock In {\em CVPR Workshops}, pages 2472--2479, 2021.

\bibitem{bell2019blind}
S.~Bell-Kligler, A.~Shocher, and M.~Irani.
\newblock Blind super-resolution kernel estimation using an internal-gan.
\newblock In {\em NIPS}, pages 284--293, 2019.

\bibitem{bevilacqua2012low}
M.~Bevilacqua, A.~Roumy, C.~Guillemot, and M.~L. Alberi-Morel.
\newblock Low-complexity single-image super-resolution based on nonnegative
  neighbor embedding.
\newblock 2012.

\bibitem{cai2020residual}
J.~Cai, Z.~Meng, and C.~M. Ho.
\newblock Residual channel attention generative adversarial network for image
  super-resolution and noise reduction.
\newblock In {\em CVPR Workshops}, 2020.

\bibitem{dong2014learning}
C.~Dong, C.~C. Loy, K.~He, and X.~Tang.
\newblock Learning a deep convolutional network for image super-resolution.
\newblock In {\em ECCV}, pages 184--199. Springer, 2014.

\bibitem{dosovitskiy2016generating}
A.~Dosovitskiy and T.~Brox.
\newblock Generating images with perceptual similarity metrics based on deep
  networks.
\newblock In {\em NIPS}, pages 658--666, 2016.

\bibitem{du2021anchor}
Z.~Du, J.~Liu, J.~Tang, and G.~Wu.
\newblock Anchor-based plain net for mobile image super-resolution.
\newblock In {\em CVPR Workshops}, pages 2494--2502, 2021.

\bibitem{ducournau2016deep}
A.~Ducournau and R.~Fablet.
\newblock Deep learning for ocean remote sensing: an application of
  convolutional neural networks for super-resolution on satellite-derived sst
  data.
\newblock In {\em Pattern Recogniton in Remote Sensing Workshop}, pages 1--6.
  IEEE, 2016.

\bibitem{fritsche2019frequency}
M.~Fritsche, S.~Gu, and R.~Timofte.
\newblock Frequency separation for real-world super-resolution.
\newblock In {\em ICCV Workshops}, 2019.

\bibitem{goodfellow2014generative}
I.~Goodfellow, J.~Pouget-Abadie, J.~Mirza, B.~Xu, D.~Warde-Farley, S.~Ozair,
  A.~Courville, and Y.~Bengio.
\newblock Generative adversarial nets.
\newblock In {\em NIPS}, pages 2672--2680, 2014.

\bibitem{huang2017densely}
G.~Huang, Z.~Liu, L.~Van Der~Maaten, and K.~Weinberger.
\newblock Densely connected convolutional networks.
\newblock In {\em CVPR}, pages 4700--4708, 2017.

\bibitem{huang2015single}
J.~Huang, A.~Singh, and N.~Ahuja.
\newblock Single image super-resolution from transformed self-exemplars.
\newblock In {\em CVPR Workshops}, pages 5197--5206, 2015.

\bibitem{huang2017simultaneous}
Y.~Huang, L.~Shao, and A.~Frangi.
\newblock Simultaneous super-resolution and cross-modality synthesis of 3d
  medical images using weakly-supervised joint convolutional sparse coding.
\newblock In {\em CVPR}, pages 6070--6079, 2017.

\bibitem{ignatov2021real}
A.~Ignatov, R.~Timofte, M.~Denna, and A.~Younes.
\newblock Real-time quantized image super-resolution on mobile npus, mobile ai
  2021 challenge: Report.
\newblock In {\em CVPR Workshops}, pages 2525--2534, 2021.

\bibitem{ji2020real}
X.~Ji, Y.~Cao, Y.~Tai, C.~Wang, J.~Li, and F.~Huang.
\newblock Real-world super-resolution via kernel estimation and noise
  injection.
\newblock In {\em ICCV Workshops}, pages 466--467, 2020.

\bibitem{johnson2016perceptual}
J.~Johnson, A.~Alahi, and L.~Fei-Fei.
\newblock Perceptual losses for real-time style transfer and super-resolution.
\newblock In {\em ECCV}, pages 694--711. Springer, 2016.

\bibitem{kim2016accurate}
J.~Kim, J.~Kwon~Lee, and K.~Mu~Lee.
\newblock Accurate image super-resolution using very deep convolutional
  networks.
\newblock In {\em CVPR}, pages 1646--1654, 2016.

\bibitem{kim2016deeply}
J.~Kim, J.~Kwon~Lee, and K.~Mu~Lee.
\newblock Deeply-recursive convolutional network for image super-resolution.
\newblock In {\em CVPR}, pages 1637--1645, 2016.

\bibitem{ledig2017photo}
C.~Ledig, L.~Theis, F.~Husz{\'a}r, J.~Caballero, A.~Cunningham, A.~Acosta,
  A.~Aitken, A.~Tejani, J.~Totz, Z.~Wang, et~al.
\newblock Photo-realistic single image super-resolution using a generative
  adversarial network.
\newblock In {\em CVPR}, pages 4681--4690, 2017.

\bibitem{lim2017enhanced}
B.~Lim, S.~Son, H.~Kim, S.~Nah, and K.~Mu~Lee.
\newblock Enhanced deep residual networks for single image super-resolution.
\newblock In {\em CVPR Workshops}, pages 136--144, 2017.

\bibitem{NTIRE2020RWSRchallenge}
A.~Lugmayr, M.~Danelljan, R.~Timofte, N.~Ahn, J.~Cai, et~al.
\newblock {NTIRE} 2020 challenge on real-world image super-resolution: Methods
  and results.
\newblock In {\em CVPR Workshops}, 2020.

\bibitem{AIM2019RWSRchallenge}
A.~Lugmayr, M.~Danelljan, R.~Timofte, et~al.
\newblock {AIM} 2019 challenge on real-world image super-resolution: Methods
  and results.
\newblock In {\em ICCV Workshops}, 2019.

\bibitem{martin2001database}
D.~Martin, C.~Fowlkes, D.~Tal, and J.~Malik.
\newblock A database of human segmented natural images and its application to
  evaluating segmentation algorithms and measuring ecological statistics.
\newblock In {\em ICCV Workshops}, volume~2, pages 416--423. IEEE, 2001.

\bibitem{matsui2017sketch}
Y.~Matsui, K.~Ito, Y.~Aramaki, A.~Fujimoto, T.~Ogawa, T.~Yamasaki, and
  K.~Aizawa.
\newblock Sketch-based manga retrieval using manga109 dataset.
\newblock {\em Multimedia Tools and Applications}, 76(20):21811--21838, 2017.

\bibitem{sajjadi2017enhancenet}
M.~S. Sajjadi, B.~Scholkopf, and M.~Hirsch.
\newblock Enhancenet: Single image super-resolution through automated texture
  synthesis.
\newblock In {\em ICCV}, pages 4491--4500, 2017.

\bibitem{tai2017image}
Y.~Tai, J.~Yang, and X.~Liu.
\newblock Image super-resolution via deep recursive residual network.
\newblock In {\em CVPR}, pages 3147--3155, 2017.

\bibitem{tong2017image}
T.~Tong, G.~Li, X.~Liu, and Q.~Gao.
\newblock Image super-resolution using dense skip connections.
\newblock In {\em CVPR}, pages 4799--4807, 2017.

\bibitem{wang2018esrgan}
X.~Wang, K.~Yu, S.~Wu, J.~Gu, Y.~Liu, C.~Dong, Y.~Qiao, and C.~Change~Loy.
\newblock Esrgan: Enhanced super-resolution generative adversarial networks.
\newblock In {\em ECCV}, pages 0--0, 2018.

\bibitem{wang2004image}
Z.~Wang, A.~Bovik, H.~Sheikh, and E.~Simoncelli.
\newblock Image quality assessment: from error visibility to structural
  similarity.
\newblock {\em IEEE T-IP}, 13(4):600--612, 2004.

\bibitem{wang2020deep}
Z.~Wang, J.~Chen, and S.~Hoi.
\newblock Deep learning for image super-resolution: A survey.
\newblock {\em IEEE T-PAMI}, 2020.

\bibitem{yang2019deep}
W.~Yang, X.~Zhang, Y.~Tian, W.~Wang, J.~Xue, and Q.~Liao.
\newblock Deep learning for single image super-resolution: A brief review.
\newblock {\em J. Multimedia}, 21(12):3106--3121, 2019.

\bibitem{zeyde2010single}
R.~Zeyde, M.~Elad, and M.~Protter.
\newblock On single image scale-up using sparse-representations.
\newblock In {\em International conference on curves and surfaces}, pages
  711--730. Springer, 2010.

\bibitem{zhang2020aim}
K.~Zhang, M.~Danelljan, Y.~Li, R.~Timofte, J.~Cai, et~al.
\newblock Aim 2020 challenge on efficient super-resolution: Methods and
  results.
\newblock In {\em ECCV Workshops}, pages 5--40. Springer, 2020.

\bibitem{zhang2010super}
L.~Zhang, H.~Zhang, H.~Shen, and P.~Li.
\newblock A super-resolution reconstruction algorithm for surveillance images.
\newblock {\em Signal Processing}, 90(3):848--859, 2010.

\bibitem{zhang2018residual}
Y.~Zhang, Y.~Tian, Y.~Kong, B.~Zhong, and Y.~Fu.
\newblock Residual dense network for image super-resolution.
\newblock In {\em CVPR}, pages 2472--2481, 2018.

\end{thebibliography}
}

\end{document}